# *The Munich Rent Advisor: A Success for Logic Programming on the Internet*

THOM FRÜHWIRTH and SLIM ABDENNADHER

*Ludwig-Maximilians-Universität München (LMU), Institut für Informatik,*
*Oettingenstrasse 67, D-80538 Munich, Germany*
(*e-mail:* {`Thom.Fruehwirth,Slim.Abdennadher`}`@informatik.uni-muenchen.de`)

## Abstract

Most cities in Germany regularly publish a booklet called the *Mietspiegel*. It basically contains a verbal description of an expert system. It allows the calculation of the estimated fair rent for a flat. By hand, one may need a weekend to do so. With our computerized version, the *Munich Rent Advisor*, the user just fills in a form in a few minutes and the rent is calculated immediately. We also extended the functionality and applicability of the *Mietspiegel* so that the user need not answer all questions on the form. The key to computing with partial information using high-level programming was to use constraint logic programming.

We rely on the internet, and more specifically the World Wide Web, to provide this service to a broad user group, the citizens of Munich and the people who are planning to move to Munich. To process the answers from the questionnaire and return its result, we wrote a small simple stable special-purpose web server directly in $\text{ECL}^i\text{PS}^e$. More than ten thousand people have used our service in the last three years. This article describes the experiences in implementing and using the *Munich Rent Advisor*. Our results suggests that logic programming with constraints can be an important ingredient in intelligent internet systems.

## 1 Introduction

In winter 1995/1996 we wanted to develop a prototypical intelligent internet application illustrating the power of using logic programming with constraints (CLP) (Wallace, 1996; Frühwirth and Abdennadher, 1997b; Marriott and Stuckey, 1998) as an implementation language. CLP combines the advantages of two declarative paradigms: logic programming (Prolog) and constraint solving. In logic programming, problems are stated in a declarative way using rules to define relations (predicates). Problems are solved using chronological backtrack search to explore choices. In constraint solving, efficient special-purpose algorithms are employed to solve subproblems involving distinguished relations referred to as constraints, which can be considered as pieces of partial information.

The *Munich Rent Advisor* (MRA) (Frühwirth and Abdennadher, 1996; Frühwirth and Abdennadher, 1997a) is the electronic version of the *Mietspiegel* (MS) for the city of Munich (Alles and Guder, 1994). Such *Mietspiegel* are published every four years by most German cities. They are basically a written description of an expert system for estimating the maximum fair rent for a flat. These estimates are legally binding, the results can be used in court cases.

Doing it by hand, one may need a weekend to calculate the rent. Usually, the calculation is performed with a pen and pocket calculator in about half an hour by an expert from the City of Munich or from one of the renter's associations. The MRA brought the advising time down to the few minutes that the user needs to fill in the form – calculation time is negligible. The calculations are based on size, age, and location of the flat and a series of detailed questions about the flat and the house it is in. Some of these questions are hard to answer. However, in order to be able to calculate the rent estimate by hand, all questions must be answered.

The MRA extended the functionality and applicability of the MS so that the user need not answer all questions of the form. The user may not want to give information away, or he does not care about the question or know the answer. He may even submit a blank form. The MRA will give an estimate of the rent as an interval as tight as possible. So the MRA now can be used not only for calculating the estimated fair rent of a flat but also for helping house hunters which have a vague idea of the kind of the flat they plan to rent and are interested in the rent they have to meet.

Our approach was to first implement the tables, rules, and formulas of the paper version with high-level declarative programming in $ECL^iPS^e$ Prolog (Brisset et. al., 1995), as if the provided data was precise. Because of the declarativity of Prolog it was easy to express the contents of the MS. Then we added constraints to capture the imprecision due to the statistical method and incompleteness in case the user gives no or partial answers. Finally, we considered the formulas of the rent calculation as constraints that refine the rent estimate by propagation from the constrained input variables. The constraints are handled by a constraint solver written in Constraint Handling Rules (Frühwirth, 1998).

This implementation approach illustrates the ease of high-level modeling that is possible with constraints logic programming (Wallace, 1996) and that supports maintenance and modification of the resulting program. This is crucial, since every city and every new version of the *Mietspiegel* comes with different tables and rules.

*The Munich Rent Advisor (MRA)* is accessible through the internet, more specifically through World Wide Web (WWW). We chose not to rely on advanced developments like Java applets or frames so that the service is accessible for any internet user. To process the answers from the questionnaire and return its result, we wrote a simple stable special-purpose web server directly in $ECL^iPS^e$ using its socket interface for internet communication.

It took about two man weeks to write the calculation part and one week to debug it. The internet user interface took one man month. We think that the coding of the calculation part would have dominated the implementation effort if a conventional programming language had been used.

In the last three years, more than ten thousand people have used our MRA service on the World Wide Web (WWW). It is one of the winners of the best application prize of the JFPLC'96 conference in Clermont-Ferrand, France, and was presented at the Systems'97 Computer Show in Munich. This article is an extended and substantially revised version of the paper (Frühwirth and Abdennadher, 1996).

The paper is organized as follows. The next section introduces the *Mietspiegel*. Section 3 describes the World Wide Web Front End. Section 4 presents the Web Server in ECL$^i$PS$^e$. Section 5 presents the implementation of the calculation part of the Munich Rent Advisor. Section 6 explains that the MRA can be modified and adapted within minutes by cloning. Section 7 presents some user statistics derived from randomly logged requests. Finally, we conclude with a summary and directions for future work.

## 2 The *Mietspiegel*

The *Mietspiegel* (MS) is published every four years by the housing group of the department for social issues of the city government of Munich after negotiations with renter's and landlord's associations and lawyers. The *Infratest Sozialforschung* Institute in Munich together with the Institute for Housing and Environment in Darmstadt conducted about 7000 interviews to obtain the sample data which was then used to build the statistical model at the department of statistics of the Ludwig-Maximilians University in Munich (Alles and Guder, 1994).

For the MS, the complex data sets derived from the interview have been reduced and simplified so that an average person could calculate the estimated rent. As we have pointed out in the introduction, the MS calculation is still too complicated to be used by everybody. In addition, the paper version ignores the inherent imprecision of the statistical model. The imprecision is basically the standard deviation obtained in the statistical model. Therefore it is higher for rare kinds of flats (very small, big or very old, new etc.). On average, the imprecision deviation amounts to about ±10% of the estimated rent.

The scheme for calculating the rent estimate is roughly as follows:

$$
\begin{aligned}
Estimated\ Rent \ =\ & Size * Basic\ Rent\ per\ SquareMeter \\
* \ & (Sum\ of\ Deviations\ as\ Percentage\ + 100) * 0,01 \\
* \ & (Imprecision\ Deviation\ Percentage\ + 100) * 0,01 \\
+ \ & Fixed\ Costs
\end{aligned}
$$

The calculation starts with the average rent per square meter taken from a table with about 200 entries. The deviations from the average rent are computed from the answers regarding the size, location, features of the flat, as well as age and state of the house. There are six yes-no questions about features of the house concerning e.g. number of floors, optical impression, lift, etc., and 13 yes-no questions about features of the flat concerning e.g. central heating, separate shower, dish-washer, etc. The answers to these questions combined with the age of the house yield the deviations from the average rent. The overall deviation may be up to ±60%.

Finally one has to add fixed costs, such as community taxes, fees for garbage collection, house cleaning, or cable TV. Part of them may be included in the rent, part of them not, part of them may not apply. Usually, the user will just ignore this section because of too much detail. Thus a range from minimal to maximal fixed costs will be added to the estimated rent.

## 3  The World Wide Web Front End

In our computerized version, *The Munich Rent Advisor (MRA)*, we rely on the internet, and more specifically the World Wide Web (WWW). We programmed in HTML version 3.0, because it is considered the current standard. We chose not to rely on advanced developments like Java applets or frames so that the service is accessible for any internet user.

| I. Basic Questions | | |
|---|---|---|
| What is the size of your flat (in squaremeters)? | at least 76 m² | not more than 85 m² |
| How many rooms has your flat? | at least 3 room(s) | not more than 4 room(s) |
| In which year was your house built? | between 1975 | and 1978 |

| II. District | |
|---|---|
| Please choose the district you live in from the list right next. | Bogenhausen |

| III. Questions about the House | | | |
|---|---|---|---|
| Do you live in the back premises? | ◆Yes | ◇No | ◇Don't know |
| Would you say your house looks good?<br>E.g. old-fashioned windows, fancy balconies. | ◇Yes | ◆No | ◇Don't know |

Fig. 1. Part of the Form

For users who are not familiar with the *Mietspiegel* (MS) we have created several web pages of background information in German. This is basically plain text with the possibility to go backward, upward and forward in the text. Furthermore, there are the additional possibilities a hypertext document provides: cross-references, links to the city of Munich and renter's associations, and to the institutions involved in preparing the MS.

All relevant information for calculating an estimated rent will be collected in the questionnaire. MRA users need to fill in only what they know and what they care about. All answers are optional. There are only four questions requiring numeric inputs, where it is possible to give a range (editable fields) and one question about location requiring a search in a list of districts (pull-down menu). The remaining questions are multiple choices, where the only possible answers are *Yes, No* and, in addition, *Don't know/care* (buttons). Optional detailed questions are provided to calculate the fixed costs where numeric input can be given. This form is divided in four sections, basic questions, questions about the house, questions about the flat itself (Figure 1), and questions about the fixed costs of the flat. These questions were sorted by importance of the answer to estimate the rent. Questions at the beginning of a section have more influence on the result than questions at the end of a section.

To fit the form on one web page we had to create a long document that consequently needs a lot of scrolling. We have experimented with internal anchors and links but users found this too complicated. Furthermore, collecting data from different pages in the server would have been too error-prone:

- The same page could be sent to the server more than once.
- Some forms might not be sent at all.
- The server has to wait for the missing forms and thus has to buffer the data.

## 4 ECL$^i$PS$^e$ as Web Server

To process the answers from the questionnaire and return its result, we wrote a simple stable special-purpose web server directly in ECL$^i$PS$^e$. This is opposed to the standard approach where for each user request a script is executed (usually written in Perl) via the CGI interface (or using the Unix inetd service). Since starting up ECL$^i$PS$^e$ (and ECL$^i$PS$^e$ saved states) takes up to a second and considerable memory, it would not have been feasible to start a new ECL$^i$PS$^e$ process with each user request. We also did not want to struggle with CGI scripts – but this problem seems to be solved in the meantime (Naish, 1999; Cabeza and Hermenegildo, 1996). It was more natural that ECL$^i$PS$^e$ is constantly running and listening to the port waiting for the next user request. Moreover, this avoids the overhead of using standard Perl scripts to communicate the data between a standard web server and the ECL$^i$PS$^e$ process. The disadvantage is that the server is not concurrent (multi-user). However, since it takes considerably less than a second to serve a user request, we did not encounter problems in practice. Moreover, the server proved to be quite robust. Clearly writing a server is only feasible for special cases.

In what follows, we assume some familiarity with Prolog or similar languages with constraints (Frühwirth and Abdennadher, 1997b; Marriott and Stuckey, 1998). ECL$^i$PS$^e$ 3.5.x offers a number of built-in predicates for TCP/IP based communication on the internet. The complete socket library (Internet Protocol Suite) as used under SUN-OS is available. Therefore the basic code of a web server in ECL$^i$PS$^e$ is just:

```
% top level
go :-
    writeln('Starting MRA Server'),
    % connecting to the internet via socket library
    socket(internet,stream,Socket),
    bind(Socket, _Hostname/4322), % 4322 is a port number
    listen(Socket,1),
    loop(Socket),
    close(Socket).
```

```
% get user requests
loop(S) :-
    accept(S,_,IOStream),  % get the request
    process(IOStream),     % process the request
    close(IOStream),       % done - served the request
    loop(S).               % go for next request
```

When the user presses the submit button of the form, the connection will be established (`accept`) and the data will be sent to the server. During calculation the connection is in stand-by mode until the result is sent back on the same stream (`IOStream`).

The processing of a user request amounts to the following code (which is similar to what one finds with other approaches like (Naish, 1999)):

```
process(IOStream) :-
    readin_request(IOStream,RequestString),
    parse(RequestString,InputVarList),
    compute(InputVarList,OutputVarList)
    ->
    sendback_result(IOStream,OutputVarList)
    ;
    sendback_error(IOStream).
```

In the Prolog predicate readin_request the data is received as an HTML document page (the `RequestString`) consisting of a header and a body (similar to an e-mail message):

```
POST / HTTP/1.0
...
Content-type: application/x-www-form-urlencoded
Content-length: 654

Language=English&M2_min=22&M2_max=160&ZI_min=1&ZI_max=9&
BJ_min=1800&BJ_max=1992&District=Schwabing&BackPremises=%3F&
...
```

The message body contains the answers of the user as `fieldname=value` entries separated by `&`. To make the server robust, there are timeouts and readin_request will fail as soon as the received message does not have the expected format. Therefore it is almost impossible to break the server.

The difficulty in parsing (predicate `parse`) is that different browsers may use different syntactic conventions and different encodings for characters. Using *Definite Clause Grammars (DCGs)*, that are available in Prolog, greatly simplified this task. The field names of the form are associated with Prolog variables which will be used to constrain the input variables:

```
Language='English', M2_min=22, M2_max=160,...
District='Schwabing', BackPremises='?',...
```

Then with the predicate `compute` the estimated rent is computed from the constrained input variables (see next section). This takes far less than a second. This means that the Web user gets the reply as fast as loading a medium-sized text-only web page and therefore the pure calculation time is negligible.

Finally, from the `OutputVarList` containing the constrained output variables (the main one being the estimated rent), a web page is assembled and sent back to the user (`sendback_result`) (Figure 2).

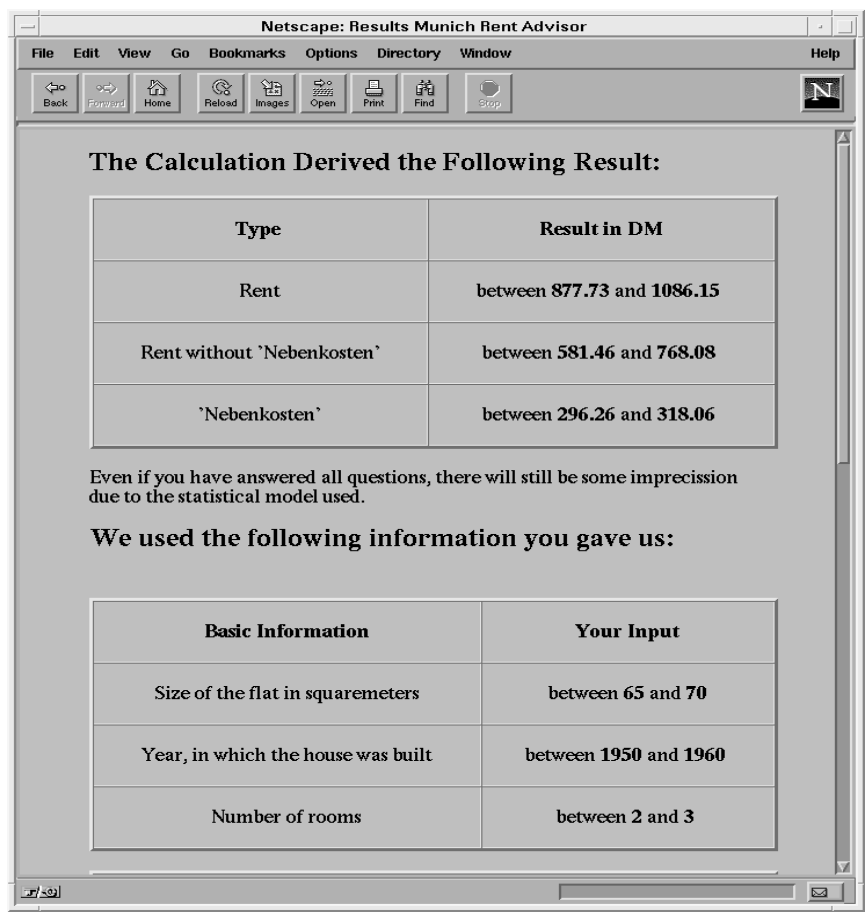

Fig. 2. Partial result of a sample query

If any failure occurs during the processing (e.g., connection times out, parsing not possible due to wrong user input in editable fields, computation unexpectedly fails), a generic error message with some hints about typical errors is sent back to the user (`sendback_error`). Of course, this primitive error handling is only sufficient for a prototype. Ironically, the success of the MRA means that there is not enough pressure to improve on it.

## 5 Implementing the Rent Calculation

Our approach was first to implement the tables, rules, and formulas of the *Mietspiegel* in ECL$^i$PS$^e$ Prolog (Brisset et. al., 1995), as if the provided data was precise and completely known. Because of the declarativity of Prolog it was easy to express the contents of the MS. Then we added interval constraints to capture the imprecision due to the statistical approach and incompleteness due to partial user answers. Finally, we considered the formulas of the rent calculation as constraints that refine the rent estimate by propagation from the constraints on the input variables which are constrained due to partial answers of the user.

While it would have been difficult to achieve exactly the required constraint propagation with a given, built-in black-box constraints solver, it was relatively straightforward using Constraint Handling Rules (CHR) (Frühwirth, 1998). It sufficed to modify an existing constraint solver written in CHR that is part of the CHR ECL$^i$PS$^e$ library. The solver takes just a few pages of code as will be exemplified in the following.

### 5.1 Interval Arithmetic

In the MRA, dealing with imprecise numerical information involves non-linear arithmetic computations with intervals (Cleary, 1987; Davis, 1987; Benhamou, 1995).

All variables are initialized to their allowed range (e.g., the flats covered by the MS are between 22 and 160 square meters, i.e. `Size::22:160`). The fieldname variables are used to constrain the input variables of the MS:

```
FlatSize::M2_min:M2_max
```

We could have used a CLP language with off-the-shelf interval arithmetic as e.g. available in CLP(BNR) (Benhamou, 1995) to express the required interval constraints. However, it would have been quite difficult to tailor the amount and direction of constraint propagation to the needs of the application at hand. Without such tailoring, the performance suffers, since the general algorithms have higher complexity than the simple forward propagation that is all that we need. In our case it sufficed to modify an existing constraint solver written in CHR that is part of the CHR library of ECL$^i$PS$^e$.

CHR is a high-level language extension to write constraint systems. Basically, CHR consists of multi-headed guarded rules. There are two kinds of rules: Simplification rules rewrite constraints to simpler constraints while preserving logical equivalence (e.g., `X>Y,Y>X <=> false`). Propagation rules add new constraints which are logically redundant but may cause further simplification (e.g., `X>Y,Y>Z ==> X>Z`). Repeatedly applying the rules incrementally solves constraints (e.g., `A>B,B>C,C>A` leads to `false`). With multiple heads and propagation rules, CHR provides two features which are essential for implementing non-trivial constraint reasoning.

The original CHR solver for finite domains already includes the basic interval constraint that restricts a variable to be in an interval between the numbers `Max` and `Min`, written `X::Min:Max`, and simple equations between two variables or numbers, e.g. `X=Y, X<Y, X=<Y,...` The solver contains rules like:

```
X::A:A <=> X=A.

X::A:B, X::C:D <=> Min is max(A,C), Max is min(B,D), X::Min:Max.

X=<Y, X::A:B, Y::C:D ==> X::A:D, Y::A:D.
```

A rule of the form `Head <=> Guard | Body` (where the guard is optional) is used to simplify the head constraints into the body, provided the guard is satisfied. Similarly, a rule of the form `Head ==> Guard | Body` is used to propagate from the head by adding the body. The first rule removes a domain that consists only of a single value and unifies its variable with that value. The second rule intersects two intervals for the same variable, thus tightening the interval. The interval constraints `X::A:B` and `X::C:D` are simplified into (replaced by) the single constraint `X::max(A,C):min(B,D)`. The last rule propagates new intervals for the variables `X` and `Y` when `X=<Y`. The constraints from the left hand side of the rule are kept in this case, the constraints from the right hand side are added. For example, from the constraints `U::2:3, V::1:2, U=<V` we get `U=2,V=2` by applying the above three rules from bottom to top.

We extended this solver by allowing linear and non-linear equations reducing to the normal forms

$$c_0 + c_1 * x_1 + c_2 * x_2 + ... + c_n * x_n = y \text{ and } c * x_1 * x_2 * ... * x_n = y$$

where the $c_i$ and $c$ are numbers and the $x_i$ and $y$ are different variables and $n \geq 0$. These equations are needed to express the formulas appearing in the MS.

The implementation for linear equations is straightforward. In the solver, the equation $c_0 + c_1 * x_1 + c_2 * x_2 + ... + c_n * x_n = y$ is represented by the constraint `sum(C0:C0+C1*X1+C2*X2+...+Cn*Xn+0=Y)`. The constant $c_0$ is replaced by the interval `C0:C0` and the summand `0` is introduced to end the summation. A constraint of the form `sum(Min:Max+Rest=Y)` means that the interval `Min:Max` plus the sum of the polynomial `Rest` gives an interval for the variable `Y`. The rules below define forward propagation: From the intervals associated with the variables `Xi` in the polynomial they compute an interval for `Y`:

```
sum(Min:Max+C*X+Rest=Y), X::A:B ==>
    NewMin is Min + min(C*A,C*B),
    NewMax is Max + max(C*A,C*B),
    sum(NewMin:NewMax+Rest=Y).

sum(Min:Max+0=Y) <=> Y::Min:Max.
```

The first rule reads: If we have the constraint `sum(Min:Max+C*X+Rest=Y)` and we know that the variable `X` is between `A` and `B` by constraint `X::A:B`, then `C*X` is between `min(C*A,C*B)` and `max(C*A,C*B)`. We can remove `C*X` from the sum and replace it by this interval. Added to the already existing interval `Min:Max` this enables us to conclude (propagate) the new constraint `sum(NewMin:NewMax+Rest=Y)`.

After we have eliminated all variables this way, we are left with `sum(Min:Max+0=Y)`, which means `Y::Min:Max`, as is expressed by the second rule.

Since we do not need backpropagation in our application, these two rules suffice. The implementation for non-linear equations is analogous. $c * x_1 * x_2 * ... * x_n = y$ is represented by `mlt(C:C*X1*X2*...*Xn*1=Y)`.

```
mlt(Min:Max*X*Rest=Y), X::A:B ==>
    NewMin is min(Min*A,Max*B,Max*A,Min*B),
    NewMax is max(Min*A,Max*B,Max*A,Min*B),
    mlt(NewMin:NewMax*Rest=Y).

mlt(Min:Max*1=Y) <=> Y::Min:Max.
```

### 5.2 Deductive Database for Tables

The *Mietspiegel* contains several tables that relate features of the flat to changes of the estimated rent. For example, the rent depends on the age of the flat and its number of rooms. The table to describe this function as found in the MS is:

| Year of construction | 1 room | 2–3 rooms | $\geq 4$ rooms |
|---|---|---|---|
| ⋮ | ⋮ | ⋮ | ⋮ |
| 1966–1977 | -3.5 | -2.0 | -3.0 |
| 1978–1983 | 2.0 | 10.0 | 3.0 |
| 1984–1986 | 6.0 | 18.0 | 7.0 |
| ⋮ | ⋮ | ⋮ | ⋮ |

The implementation uses a simple *constraint database* (Kuper and Wallace, 1995) with intervals. In Prolog, the table is easily represented declaratively as a list of facts of the form `table(YearInterval,RoomsInterval,Percentage)`. Such a shorthand maintains readability and is compact:

```
  ...
  table(1966:1977, 1:1, -3.5).
  table(1966:1977, 2:3, -2.0).
  ...
  table(1984:1986, 2:3, 18.0).
  table(1984:1986, 4:9,  7.0).
  ...
```

The table facts are translated at compile time by macro expansion into rules to make the interval constraints explicit

```
  ...
  table(Year, Rooms, -3.5) :-
        Year::1966:1977, Rooms::1:1.
  table(Year, Rooms, -2.0) :-
```

```
        Year::1966:1977, Rooms::2:3.
...
```

The Prolog query `Year=1980, Rooms=2, table(Year,Rooms,Percentage)`, for example, yields as an answer `Percentage=10.0`.

In the general case, when we use the tables with constrained variables in the queries for `Year` and `Rooms`, we are only interested in the smallest interval that contains all the answers, not in all answers as such. For example, with the constraints `Year::1980:1985, Rooms::1:3` we want the single answer `Percentage::2.0:18.0` and not the multiple answers `Percentage=2.0, Percentage=10.0,...` This means that we have to collect all the answers and compute minima and maxima of the percentages returned to find the smallest interval that contains all answers. This *meta-programming* task can be easily accomplished by a built-in predicate of Prolog, `setof(Variable,Query,List)` that collects all bindings of the variable in all the answers to the query in a sorted list:

```
setof(Percentage,Year^Rooms^table(Year,Rooms,Percentage),List),
first(List,Min), last(List,Max),
Percentage::Min:Max.
```

A similar procedure was used for all tables. The running time is satisfactory for tables with a few hundred constrained tuples.

## 6 Cloning

An advantage of internet applications using Prolog with constraints is that they can be modified and adapted within minutes, since the MRA can be *cloned*: Any part of the form used as the interface to this application may be reused in another web page simply by cut-and-paste.

One may drop questions, one may set default values, one may fix the answer to questions and hide them from the user in order to specialize the application. Furthermore, everything on the web page can be rearranged at will, only the form declaration has to be kept as is.

The resulting web page will still work, i.e. produce a result page when submitted, since the missing information is dealt with by constraints. For example, the minimal clone is simply:

```
<HTML>
<HEAD>
<TITLE>Minimal Clone</TITLE>
</HEAD>
<BODY>
<FORM METHOD="POST"
      ACTION="http://sol.pst.informatik.uni-muenchen.de:4322">
<INPUT TYPE="submit" VALUE="Submit">
</BODY>
</HTML>
```

The form just consists of a submit button. When pressed, it will just return the smallest and highest allowed rent of flats up to $160m^2$ in Munich.

A more realistic clone simplifies the interface to the basic questions as can be seen on top of Figure 1.

```
...
<BODY>
<FORM METHOD="POST"
      ACTION="http://sol.pst.informatik.uni-muenchen.de:4322">
<H1>Munich Rent Advisor - Clone</H1>
What is the size of your flat (in square meters)?<BR>
At least <INPUT MAXLENGTH=3 SIZE=3 NAME="M2_low" Value="22"> ...
<P>
How many rooms has your flat?<BR>
At least <INPUT MAXLENGTH=1 SIZE=1 NAME="ZI_low" Value="1"> ...
<P>
In which year was your house built?<BR>
Between <INPUT MAXLENGTH=4 SIZE=4 NAME="BJ_low" VALUE="1800"> ...
<P>
<INPUT TYPE="submit" VALUE="Submit">
</BODY>
</HTML>
```

## 7 User Statistics

We have logged about 7200 headers of user requests to the MRA for almost two years in the three years since February 1996, when the MRA went online. On average, there are 10 requests per day. Our findings can be summarized as follows (the figures related to the findings in this section can be found at the end of the article).

**Correct Requests** (see Fig. 3).

- From the 7188 requests received, only 1% can be attributed to trying to access the MRA web server improperly, without using the form.
- Due to timeout, 9% of the requests were cancelled.
- Due to syntax errors (typos, using floats instead of integers, wrong intervals), another 12% were cancelled.
- Thus, only 4 in 5 requests (78%) were in time, correct and lead to a rent estimate.

The large number of syntax errors can be attributed to users that did not read the instructions carefully or had typing problems. Most of these errors can be caught before the form is sent to the server. We have implemented such a version of the MRA in a recent student project (Herzog, 1998): JavaScript is used to provide help texts and syntax checks for each input field as soon as the user enters something.

Less than 2% of the requests did not come from the original form but from a version that was stored locally with the user. The German version of the MRA was used almost all the time, the English version accounts only for 6% of the requests.

**User Origin**. If the user establishes a connection to a web server by sending the contents of a form as a request, he usually gives away the symbolic internet address of his machine, e.g. `borabora.pms.informatik.uni-muenchen.de`. In Fig. 3, header field Accept-from, we give some statistics about the user origin. In the figure, each of the addresses subsumed in entries named *Other* have considerably less than 1% contribution each.

- Only 1 in 7 requests were anonymous.
- 2 in 3 requests came from German domains (`.de`).
- In Germany, not surprisingly, most requests came form local, Munich universities (Uni) and large IT and car companies (Com). These two groups contribute each about one fifth to the overall requests. Many requests also came from users of large internet providers (Pro) (7%). Together, these three groups of frequent requests from the same domain make up about half of all requests.

**User Software**. Each browser also tells its name and the operating system it is running on in the header (see Fig. 3, User-Agent). The figure shows that Unix-based operating systems make up for one third. The majority goes to Windows (slightly over 60%), and there are some other operating systems (less than 5%). Both Netscape and Explorer browsers call themselves *Mozilla*, however the latter adds the qualifier *Compatible*. Netscape dominates on both Unix- and Windows-based machines, overall 80%. Explorer has 11%, the rest (9%) is shared by other browsers.

**Access Times** (see Fig. 4). We analyzed user access times per month, weekday, and hour. Since we did not cover all months in all years logged, we had to extrapolate some figures to get figures for complete months, resulting in an overhead of about 10% over the actual number of requests (7890 instead of 7073).

- The monthly figures are somewhat irregular, with a low in May that we cannot explain. The high in November (and December) maybe comes from the fact that rents are usually raised at the end of the year. The high in February definitely comes from 1996, when the MRA was introduced and featured in the media.
- The weekday figures strictly decrease from Monday to Sunday, with little activity on the weekend. Almost a quarter of all requests happen on Monday, only 7% on Saturday.
- The figures for the hours show that the MRA is used mostly during working time (in the 6 hours from 11 am to 5 pm), mainly around lunch break. Not surprisingly, the figures for the 6 hours from 1 am to 7 am are extremely low (about 2% of all requests).

## 8  Conclusions

The MRA indicates that logic programming with constraints can be essential for intelligent internet applications for several reasons.

- Logic programming languages have declarative rules and powerful deductive database facilities already built-in, that are needed to encode expert knowledge.
- Such a high-level state-of-the-art approach also means that a program can be easily written, maintained, and modified (Wallace, 1996). For the MRA, ease of modification is crucial, since every city and every new version of the *Mietspiegel* comes with different tables and rules.
- Constraint logic programming languages can deal with imprecise knowledge and partial information that characterizes communication on the internet in an elegant, correct, and efficient way (Frühwirth et. al., 1997).

One direction for future work is to integrate integrity constraints (e.g., if a house is built after 1949, its flats have a bathroom) that have been directly derived from the statistical raw data of the Mietspiegel. The other direction is to create electronic versions of the Mietspiegel for more cities. Many of the now 500 Mietspiegel of Germany are currently available as tables and text on the internet. To facilitate their processing we already have developed a tool that can automatically generate forms, with help texts and syntax checks built in, together with their handlers (Herzog, 1998).

The Munich Rent Advisor Home Page is at
http://www.informatik.uni-muenchen.de/~fruehwir/miet-demo.html

**Acknowledgements**. We would like to thank Peter Blenninger who implemented a first prototype of the MRA. We are also grateful to the City of Munich for letting us use their Mietspiegel data and to Norbert Eisinger and Tim Geisler for comments and proof-reading.

| Fieldname | Value | Frequency |
|---|---|---|
| Requests |  | 7188 |
|  | wrong request | 70 |
|  | timeout header | 316 |
|  | timeout body | 327 |
|  | syntax error | 864 |
|  | correct requests | 5611 |
| Accept-from: |  | 7073 |
|  | .de | 4667 |
|  | uni-muenchen (Uni) | 815 |
|  | lrz-muenchen (Uni) | 440 |
|  | sni (Com) | 421 |
|  | siemens (Com) | 364 |
|  | dtag (Com) | 360 |
|  | tu-muenchen (Uni) | 299 |
|  | t-online (Pro) | 266 |
|  | mpg (Com) | 97 |
|  | sdm (Com) | 87 |
|  | bmw (Com) | 82 |
|  | gsf (Com) | 80 |
|  | eunet (Pro) | 78 |
|  | metronet (Pro) | 73 |
|  | uunet (Pro) | 60 |
|  | Other | 1145 |
|  | .com | 731 |
|  | .net | 319 |
|  | Other | 261 |
|  | anonymous | 987 |
|  | self test | 108 |
| User-Agent: |  | 6875 |
|  | Mozilla | 6606 |
|  | Win | 3514 |
|  | Win95 | 1570 |
|  | Win16 | 933 |
|  | WinNT | 652 |
|  | Windows | 359 |
|  | X11 | 1989 |
|  | SunOS | 805 |
|  | HP-UX | 713 |
|  | Linux | 254 |
|  | Irix | 125 |
|  | Other | 102 |
|  | Compatible | 762 |
|  | MSIE 95 | 466 |
|  | MSIE NT | 114 |
|  | AOL | 78 |
|  | MSIE Win32 | 76 |
|  | Macintosh | 226 |
|  | OS/2 | 111 |
|  | Mosaic | 164 |
|  | Other | 167 |

Fig. 3. Information from the Header

The Munich Rent Advisor    17| Fieldname | Value | Frequency |
|---|---|---:|
| Accept-time: | | 7890 |
| | Jan | 386 |
| | Feb | 1023 |
| | Mar | 599 |
| | Apr | 600 |
| | May | 304 |
| | Jun | 619 |
| | Jul | 458 |
| | Aug | 481 |
| | Sep | 520 |
| | Oct | 751 |
| | Nov | 1282 |
| | Dec | 867 |
| by week-day | | 7073 |
| | Mon | 1658 |
| | Tue | 1346 |
| | Wed | 1037 |
| | Thu | 1014 |
| | Fri | 1008 |
| | Sat | 479 |
| | Sun | 531 |
| by hour | | 7073 |
| | 00 | 119 |
| | 01 | 64 |
| | 02 | 22 |
| | 03 | 20 |
| | 04 | 10 |
| | 05 | 15 |
| | 06 | 12 |
| | 07 | 111 |
| | 08 | 225 |
| | 09 | 404 |
| | 10 | 441 |
| | 11 | 607 |
| | 12 | 571 |
| | 13 | 655 |
| | 14 | 561 |
| | 15 | 665 |
| | 16 | 562 |
| | 17 | 447 |
| | 18 | 373 |
| | 19 | 289 |
| | 20 | 250 |
| | 21 | 271 |
| | 22 | 211 |
| | 23 | 168 |

Fig. 4. Temporal Information from the Header